\documentclass[letterpaper ,anonymous]{article}
\usepackage{aaai}
\usepackage{times}
\usepackage{helvet}
\usepackage{courier}
\usepackage{amsmath,amsfonts,amssymb,amsthm}
\usepackage{float, graphicx, multirow}
\usepackage{makecell} 
\usepackage{hyperref}
\usepackage{lipsum}

\usepackage[authoryear,longnamesfirst]{natbib}
\usepackage{caption}
\usepackage{subcaption}
\usepackage{physics}
\usepackage{pifont}

\newtheorem{theorem}{Theorem}

\begin{document}
%
\title{A Graph Convolution for Signed Directed Graphs}
\author{Taewook Ko$^{[1]}$ and Chong-Kwon Kim$^{[2]}$\\
$^{[1]}$Department of Computer Science and Engineering\\
Seoul National University, Seoul, Korea\\
$^{[2]}$Department of Energy AI\\
Korea Institute of Energy Technology, Naju, Korea
}

\maketitle
\begin{abstract}
\begin{quote}
A signed directed graph is a graph with sign and direction information on the edges. Even though signed directed graphs are more informative than unsigned or undirected graphs, they are more complicated to analyze and have received less research attention. This paper investigates a spectral graph convolution model to fully utilize the information embedded in signed directed edges. We propose a novel complex Hermitian adjacency matrix that encodes graph information via complex numbers. Compared to a simple connection-based adjacency matrix, the complex Hermitian can represent edge direction, sign, and connectivity via its phases and magnitudes. Then, we define a magnetic Laplacian of the proposed adjacency matrix and prove that it is positive semi-definite (PSD) for the analyses using spectral graph convolution. We perform extensive experiments on four real-world datasets. Our experiments show that the proposed scheme outperforms several state-of-the-art techniques.
\end{quote}
\end{abstract}

\noindent 
Graph convolution techniques have attracted much attention from researchers thanks to their excellent performance in graph mining tasks such as graph representation learning and node embedding\cite{xu2018powerful,grover2016node2vec,ying2018hierarchical}. Spectral convolution is one of the main streams of graph convolution research, and graph Laplacian is the key element of spectral graph analyses\cite{bruna2013spectral,defferrard2016convolutional}. Compared to unsigned undirected graphs, signed directed graphs contain complex edge information and possibly rich latent features. While the relation of a node pair in unsigned undirected graphs is binary (connected or unconnected), signed directed graphs have nine different types of node relationships by the combination of edge existence, directions, and signs. A problem in dealing with rich signed directed graphs is the difficulty of analyses rooted in the properties of their Laplacian matrices. A graph Laplacian ($\textbf{L}=\textbf{D}-\textbf{A}$) derived from an unsigned undirected graph is symmetric and positive semi-definite (PSD). However, signed or directed graphs make their Laplacian asymmetric. Their Laplacian eigenvalues are not guaranteed to be non-negative, and the eigenvectors are not unitary also. Several assumptions and approximations\cite{hammond2011wavelets,defferrard2016convolutional} proposed for spectral graph convolution are no longer valid in signed directed graphs. The analytical difficulty is one reason that left signed directed graphs inattentive, while many real-world graphs are signed and/or directed. 

The PSD property is the biggest hurdle to extending the idea of spectral convolutions to graphs with signs and/or directions. Several researchers proposed novel graph Laplacian for directed graphs. DiGCN\cite{tong2020digraph} defines the Directed graph Laplacian via a stationary distribution with teleport probability. MagNet\cite{zhang2021magnet} defines a magnetic Laplacian to represent directional edges. In contrast to directed graphs, an adequate Laplacian matrix for signed or signed directed graphs is still remained unsolved. Relinquishing spectral analyses, most prior signed graph studies employed spatial convolution approaches\cite{derr2018signed,li2020learning,jung2020signed}. They utilize the balance and status theories\cite{heider1946attitudes,holland1971transitivity} to define neighbor aggregation processes. Even though those social network theories are important paradigms in signed graph research, they sometimes fail to explain user triads and perform poorly for users with fewer or no triangles. Moreover, spatial convolutions\cite{huang2019signed,huang2021sdgnn} require high computational costs than spectral approaches.

This paper proposes a novel Laplacian matrix on which a spectral graph convolution can be applied to analyze signed directed graphs. First, we define a complex Hermitian adjacency matrix ($\textbf{H}^q$) that encodes edge types of signed directed graphs. \cite{liu2015hermitian} and \cite{guo2017hermitian} introduced complex Hermitian adjacency matrices for directed graphs to overcome the limitation of the traditional adjacency matrix. They encode directional information in a simple form of complex numbers. Extending the prior approaches, we propose a novel method that encodes both direction and sign of edges via precisely designed equations. The magnitudes of complex numbers indicate the connectivity between two nodes, and the phases indicate the sign and direction of edges between the nodes. Then, we define a magnetic Laplacian ($\textbf{L}^q$) with the complex Hermitian adjacency matrix. Magnetic Laplacian was introduced in quantum mechanics to explain the discrete Hamiltonian under magnetic flux\cite{shubin1994discrete,olgiati2017remarks}. Thanks to its Hermitian property, it has recently been leveraged for the research on directed graphs\cite{fanuel2017magnetic,furutani2019graph}. This paper extends the application of magnetic Laplacian to signed graphs. Our proposed algorithm can be applied to the analyses of unsigned or undirected graphs and enjoys wide applicability. GCN\cite{kipf2016semi} and MagNet\cite{zhang2021magnet} can be considered as special cases of our proposed scheme. We confirmed that it is the most generalized version of the traditional GCN.

We prove the PSD property of the newly proposed magnetic Laplacian. It has orthonormal complex eigenvectors and corresponding real eigenvalues. With these properties, we derive a spectral graph convolution mechanism. The overall scheme is called Signed Directed Graph Convolution Network (SD-GCN). We evaluated the SD-GCN on several real-world datasets and compared its performance with various baselines. The baselines include state-of-the-art graph convolutions for directed, signed, and signed directed graphs. Our experimental results indicate that SD-GCN outperforms the baselines in terms of link sign prediction performance in various metrics.

The contributions of this paper are as follows.
\begin{itemize}
\item This paper introduces a novel complex Hermitian adjacency matrix that encodes signed directed edges via complex numbers.
\item This paper proposes a novel magnetic Laplacian from the Hermitian matrix and proves that the magnetic Laplacian is PSD.
\item To the best of our knowledge, it is the first spectral graph convolution for signed graphs that enjoys wide applicability.
\item The extensive experimental results on several real-world datasets show that 
SD-GCN outperforms SOTA graph convolution methods.
\end{itemize}

\section{Related Work}
We introduce several important studies closely related to our proposed method.
\subsection{Graph convolution for signed directed graphs}
There are two main streams in the graph neural network research; spatial and spectral approaches. Spatial methods\cite{micheli2009neural,velivckovic2017graph,hamilton2017inductive} infuse neighbor features via message passing and local aggregation to exploit the homophilic property in graphs. There are several spatial-based models for signed directed convolution. Most of them leverage the balance theory and/or the status theory\cite{heider1946attitudes,holland1971transitivity}. The balance theory claims the balance of user triads based on the hypothesis of `A friend of my friend is my friend, an enemy of my friend is my enemy.' In the status theory, the edge direction indicates the higher rank and it assumes the relative social ranks between nodes. These theories are satisfied in the most real-world social networks. About 60-70\% of triads from real-world graphs follow the theories\cite{leskovec2010predicting,javari2017statistical} and it has been widely used in social network research for a long time. SGCN\cite{derr2018signed} proposed balanced and unbalanced paths according to the balance theory to define a novel feature aggregation method. SNEA\cite{li2020learning} extended the SGCN\cite{derr2018signed} by revisiting GAT\cite{velivckovic2017graph} to utilize the power of the attention mechanism. SGDN\cite{jung2020signed} proposed to adopt random walks for effective diffusion of hidden node features. SiGAT\cite{huang2019signed} is a generalized GAT\cite{velivckovic2017graph} leveraging triad motifs using both the balance and status theories. SDGNN\cite{huang2021sdgnn} defines four weight matrices for different edge types of signed directed graphs and proposes triad loss based on the social network theories. Even though the theories are imperative in signed directed graph research, not all the triads follow the theories. Especially, these theory based models show unsatisfactory performance for nodes with a few or no triads. Recently, to be oblivious to the social theories, ROSE\cite{javari2020rose} proposed a methodology based on the user-role property by transforming graphs into unsigned bipartite graphs.

\begin{figure}[t]
  \centering
  \includegraphics[width=\linewidth]{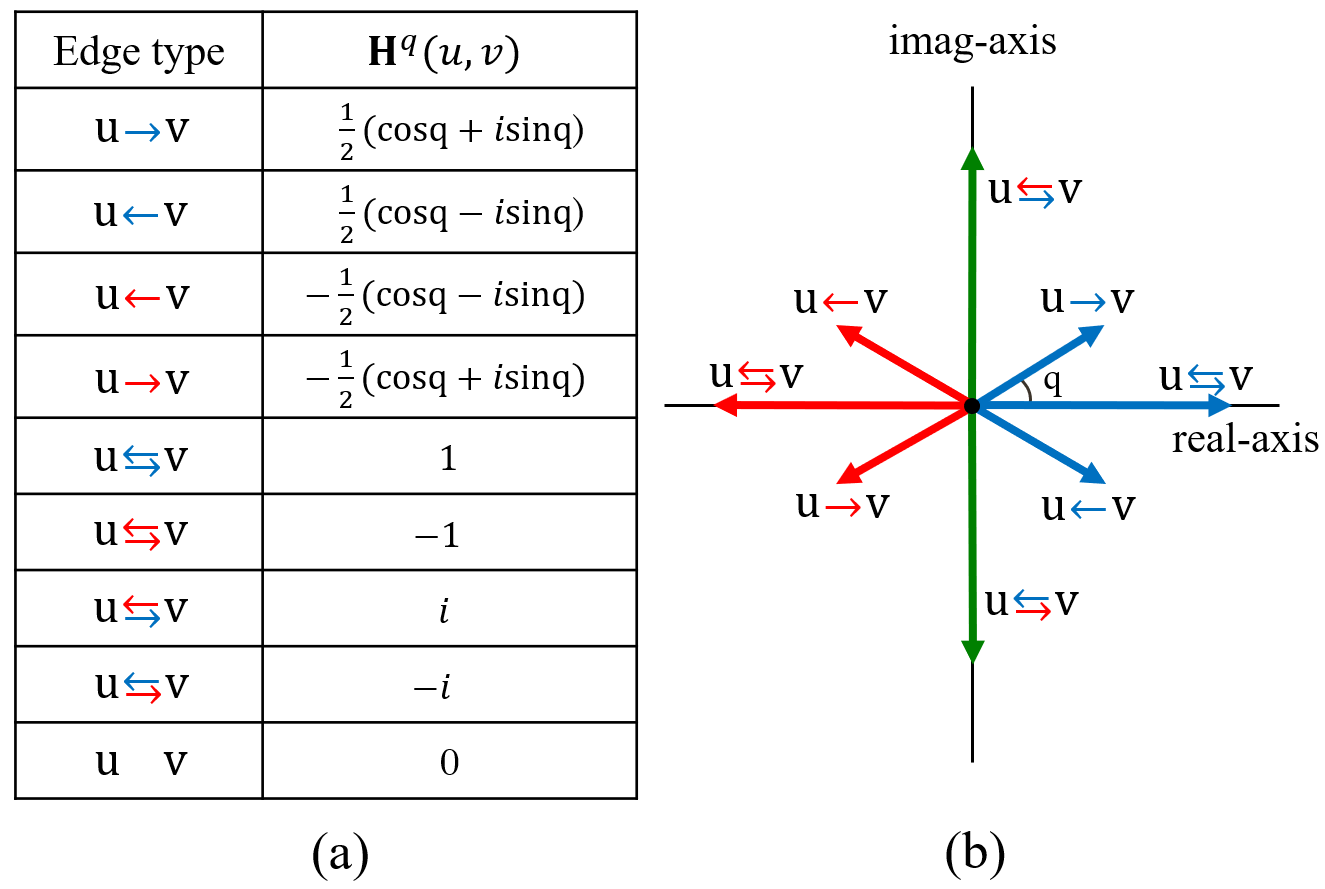}
  \caption{Graph type example. Edges with arrows are directional edges and edges with colors are signed edges. Red and blue indicate negative and positive signs, respectively. This study proposes a model for signed directed graphs and at the same time, it has wide applicability to various types of graphs.}
\end{figure}

Spectral convolution methods utilize eigendecomposition of graph Laplacian rooted on graph signal processing. They apply Fourier transform to graph signals or features which are then convoluted in the Fourier space. As we discussed in the Introduction section, traditional graph Laplacian is good for graph Fourier transform in vanilla graphs and several successful models\cite{hammond2011wavelets,defferrard2016convolutional,kipf2016semi} have been introduced. To expand their applications, some studies for directed graphs are proposed. Ma et al.\cite{ma2019spectral} proposed a Directed Laplacian matrix via transition probability and stationary distribution to make a symmetric Laplacian, whereas it requires high-density graphs. To overcome the limitation, DiGCN\cite{tong2020digraph} employs the PageRank\cite{page1999pagerank} and inception modules\cite{szegedy2016rethinking}. DGCN\cite{tong2020directed} defines three Laplacians, for symmetric, outgoing, and incoming edges, respectively, to distinguish the edge directions. MagNet\cite{zhang2021magnet} utilizes magnetic Laplacian with complex numbers. They handle the direction information via the phases of a Hermitian matrix. Though many studies provided to define a directed spectral convolution, still there is no spectral convolution for signed graphs. This paper adopts the idea of MagNet\cite{zhang2021magnet} for the analyses of signed directed graphs by defining a novel magnetic Laplacian matrix. 

\subsection{Hermitian adjacency matrix}
An ordinary adjacency matrix $\textbf{A}$ suitable for unsigned undirected graph representation is symmetric, an important property required for matrix processing. However, it is impossible to encode edges of signed and/or directed graphs maintaining the symmetry property. Liu and Li\cite{liu2015hermitian} and Guo and Mohar\cite{guo2017hermitian} introduced a Hermitian adjacency matrix to encode directed graphs with complex numbers. They also proved that its eigenvalues are real, and its eigenvectors are orthogonal and unitary. Mohar\cite{mohar2020new} later introduced another Hermitian matrix type that claimed to be more natural than the prior by analyzing various phases. Cucuringu et al.\cite{cucuringu2020hermitian} utilized the Hermitian matrix for a directed graph clustering task. Note that prior works on the Hermitian matrices are limited to directed graph analyses. In this paper, we extend the idea and propose a novel method that can represent not only the direction but also the sign information of edges. 

\subsection{Magnetic Laplacian}
The magnetic Laplacian has been studied for decades in quantum mechanics\cite{shubin1994discrete,olgiati2017remarks,lieb1993fluxes,colin2013magnetic}. It is regarded as a discrete Hamiltonian of a charged particle under the magnetic flux. 
Fanuel et al.\cite{fanuel2018magnetic} visualized that the eigenvectors are bounded in a magnetic field. Similar to Hermitian adjacency matrices that are used to replace traditional adjacency matrix, the magnetic Laplacian is introduced to overcome the limitations of the traditional Laplacian matrix $\textbf{L}$. Because the magnetic Laplacian is a Hermitian, it is widely used in directed graph mining tasks such as community detection\cite{fanuel2017magnetic}, clustering\cite{f2020characterization,cloninger2017note}, graph representation learning\cite{furutani2019graph}, and node embedding\cite{zhang2021magnet}. We extend the magnetic Laplacian to a signed graph study. To make the graph Fourier transform work on singed directed graphs, we propose a novel magnetic Laplacian that keeps all its eigenvalues real.

\begin{table}[b]
\centering
\begin{tabular}{cl}
\Xhline{2.5\arrayrulewidth}
Notation  & Description  \\ \hline
$\mathcal{G}$          &  Graph            \\
\textit{V}           & Node set      \\
\textit{E}           & Edge set      \\
\textit{\textbf{S}}           & Sign matrix   \\
\textbf{\textit{Z}}   & Node embedding matrix              \\
$\textbf{A}$, $\textbf{A}_s$ & Adjacency matrix and symmetric adjacency matrix \\
$\textbf{D}$, $\textbf{D}_s$ & Degree matrix and symmetric degree matrix \\
\textbf{L} &     Laplacian matrix \\  
$\textbf{L}^{q}$  & Magnetic Laplacian matrix with parameter $q$              \\
$\textbf{P}^{q}$ &         Phase matrix           \\
$\textbf{H}^{q}$ &           Complex Hermitian adjacency matrix          \\
$q$ & Phase control parameter \\
\textbf{X} &   Input graph signal             \\
\textbf{W, b} &   Learnable weight matrix and bias           \\
\Xhline{2.5\arrayrulewidth}
\end{tabular}%
\caption{Notations of this paper and its descriptions}
\label{tab:Dataset Statistics}
\end{table}

\section{Problem Formulation}
The problem that this paper mainly deals with is representation learning. We introduce a novel magnetic Laplacian and a Hermitian matrix that can be applied for signed directed graph analyses. Let $\mathcal{G} = (V, \textit{E}, \textit{\textbf{S}})$ be a signed directed graph where $V$ is a set of nodes, and $\textit{E}$ is a set of directed edges. As weighted graphs, we define a sign matrix $\textit{\textbf{S}}$ to denote the signs of edges; $\textit{\textbf{S}}(u,v) = 1$ or $-1$ if a directed edge from $u$ to $v$ is positive or negative, respectively. Otherwise, $\textit{\textbf{S}}(u,v) = 0$ if there is no edge. Positive and negative edges represent node relations such as like/dislike or trust/distrust in social graphs. Like most social network studies, we exclude multigraph cases. A node has a relationship to another by one of the three (none, positive, negative). For every node pair, we now define nine types of relations as shown in Fig. 2(a). The goal of this paper is to discover the latent features of each node $u \in V$ as a low-dimensional embedding vector $z_u \in \mathbb{R}^d$ for a given graph $\mathcal{G}$  as:
\begin{equation}
    f(\mathcal{G}) = Z.
\end{equation}
$Z\in \mathbb{R}^{\mid V \mid \times d}$ is a node embedding matrix with a size of $d$-dimension. Each row represents node embedding and $f$ is a learned transformation function. 

\section{Magnetic Laplacian}
For an unsigned and undirected graph, its adjacency matrix ($\textbf{A}$) and graph Laplacian ($\textbf{L}=\textbf{D}-\textbf{A}$) are symmetric. The graph Laplacian ($\textbf{L}$) has non-negative eigenvalues and its eigenvectors form orthonormal basis. GCN\cite{kipf2016semi} and Chebyshev\cite{defferrard2016convolutional} proposed spectral graph theories with these properties. However, we have an asymmetric graph Laplacian when a graph is signed and/or directed. They typically have complex eigenvalues and do not support the graph Fourier transform conditions for spectral convolution. We propose a novel magnetic Laplacian matrix that represents signed directed graphs while it satisfies the positive semi-definite property (PSD). The magnetic Laplacian has already been investigated in the field of quantum mechanics as magnetic flux\cite{lieb1993fluxes} and graph mining studies\cite{f2020characterization,furutani2019graph}. However, it is rarely discussed in graph convolution research.

\begin{figure*}[t]
  \centering
  \includegraphics[width=0.7\linewidth]{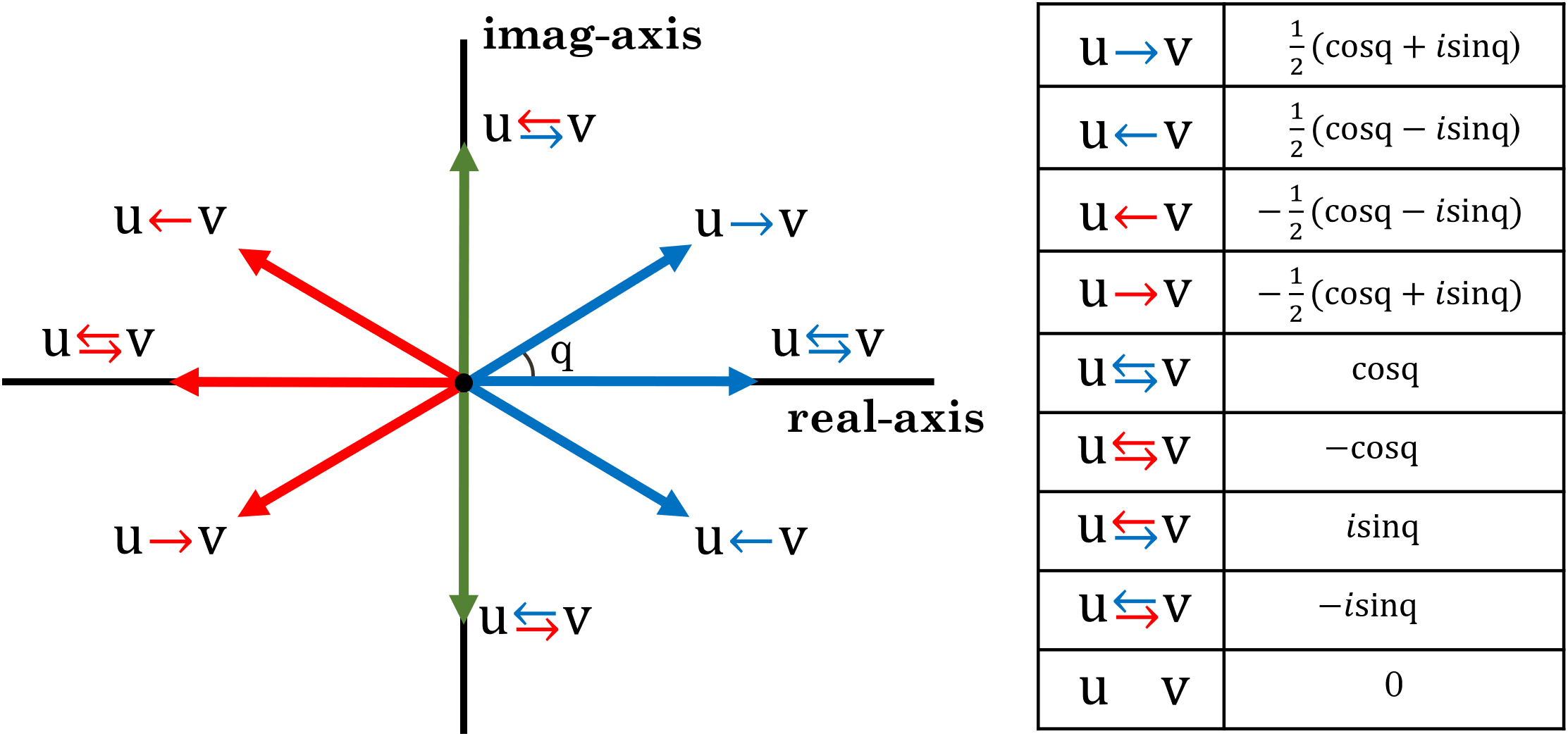}
  \caption{(a) shows the nine relation types and corresponding encoding values. Red and blue arrows indicate negative and positive edges with directions, respectively. (b) shows encoding values in the complex space. Green arrows indicate neutral edges. Origin is the case when there is no edge.} 
\end{figure*}

\subsection{Complex Hermitian Adjacency Matrix}
The traditional adjacency matrix that encodes the connectivity of nodes with 0 and 1 is symmetric when a graph is unsigned and undirected. However, it loses the symmetry property in encoding edge sign and direction information. To overcome this limitation, several studies\cite{liu2015hermitian,guo2017hermitian} proposed Hermitian adjacency matrices. They encoded the four edge types of directed graphs with \{0,\ 1,\ \textit{i},\ -\textit{i}\}. The bidirectional edges are encoded as 1, and directional edges are encoded with the two imaginary numbers. The plus and minus signs to the imaginary number distinguish the edge directions. \cite{mohar2020new,zhang2021magnet} used complex numbers with various phases to make a more explainable directed adjacency matrix. All prior studies focused on the representation of edge direction, and there is no attempt to represent the sign information. Here we define a novel Hermitian adjacency matrix as:
\begin{equation}
\mathbf{H}^q = \mathbf{A}_s \odot \mathbf{P}^{q},
\end{equation}
where $\odot$ is an element-wise multiplication operation. 
$\mathbf{A}_s := \frac{1}{2}(\mathbf{A}+ \mathbf{A^\intercal}) \subseteq V \times V $ is a symmetrized adjacency matrix. $\mathbf{A}_s(u,v)$ is one of \{0, $\frac{1}{2}$, 1\} by the edge existence between node $u$ and $v$. $\mathbf{P}^q \subseteq V \times V$ is a phase matrix with complex numbers. This phase matrix encodes edge direction and sign information as:
\begin{equation}
\textbf{P}^{q}(u,v) := \frac{\text{exp}(i\Theta^{q}_{(u,v)})\textbf{A}_{(u,v)}+ \text{exp}(i\overline{\Theta}^{q}_{(u,v)})\textbf{A}_{(v,u)}}
{\|\text{exp}(i\Theta^{q}_{(u,v)})\textbf{A}_{(u,v)}+ \text{exp}(i\overline{\Theta}^{q}_{(u,v)})\textbf{A}_{(v,u)}\|+\epsilon}.
\end{equation}
\begin{equation*}
    \Theta^{q}(u,v)= 
\begin{cases}
    \ q & \ \ \  \ \ \ \text{if} \ \ \ S(u,v)=1\\
    \ \pi+q &  \ \ \   \ \ \ \text{if} \ \ \ S(u,v)=-1\\
    \  0 & \ \ \   \ \ \  \text{if} \ \ \ S(u,v)=0\\
    \end{cases}
\end{equation*}

\begin{equation*}
    \overline{\Theta}^{q}(u,v)= 
\begin{cases}
    \ -q & \ \ \  \ \ \ \text{if} \ \ \  S(v,u)=1\\
    \ \pi-q & \ \ \  \ \ \ \text{if} \ \ \  S(v,u)=-1\\
    \  0 & \ \ \  \ \ \ \text{if}  \ \ \ S(v,u)=0\\
    \end{cases}
\end{equation*}
where $0 \leq q \leq \pi/2$. It controls the encoding phases. The effect of $q$ is described in Fig. 4 and Fig. 5, and discussed in Sec. 8.2. Note that the order of matrix index and node pair index are carefully arranged. The defined complex Hermitian adjacency matrix ($\textbf{H}^q$) encodes the nine node pair relations of signed directed graphs; in addition to the connectivity information, it embeds the direction and sign of edges. Fig. 2 illustrates the nine node pair relations and their encoded values in complex space. Each relation is unique in phase and magnitude combination. The encoding value is zero if there is no edge between the node pair. By the definition of $\textbf{H}^q$, it has complex elements and is a skew-symmetric matrix. Thus, we name it as Complex Hermitian adjacency matrix.

\subsection{Proposed Magnetic Laplacian}
We define a magnetic Laplacian with the proposed Complex Hermitian adjacency matrix. The unnormalized ($\mathbf{L}^{q}_U$) and normalized ($\mathbf{L}^{q}_N$) Laplacians are as follows:
\begin{equation}
    \mathbf{L}^{q}_U := \mathbf{D}_s-
    \mathbf{H}^{q} = \mathbf{D}_s- \mathbf{A}_s\odot \mathbf{P}^q
\end{equation}
\begin{equation}
    \mathbf{L}^{q}_N := \mathbf{I}-(\mathbf{D}^{-\frac{1}{2}}_s \mathbf{A}_s \mathbf{D}^{-\frac{1}{2}}_s)\odot \mathbf{P}^q.
\end{equation}
where $\textbf{D}_s$ is a degree matrix defined as
\begin{equation}
    \textbf{D}_s(u,v) =
\begin{cases}
    \ \sum_{w\in V}\text{A}_s(u,w) & \ \ \  \ \ \ \text{if} \ \ \ u = v     \\
    \  \ \ \ \ \ \ \ \ 0 & \ \ \   \ \ \  \text{if} \ \ \ u \neq v  \\
    \end{cases}
\end{equation}
Both Laplacian matrices are Hermitian according to the definition of $\mathbf{D}_s$, $\mathbf{A}_s$, and $\mathbf{P}^q$. We proved that these Hermitian magnetic Laplacians are PSD in Appendix B.1. The PSD property makes
the magnetic Laplacians diagonalizable with orthonormal basis of complex eigenvectors each of which has a non-negative real eigenvalue.
$\textbf{L}^{q}_N$ is spectral decomposed as
\begin{equation}
    \textbf{L}^{q}_N = \textbf{U}\Lambda\textbf{U}^{\dagger}.
\end{equation}
where $\mathbf{U}$ is a matrix composed with eigenvectors ($\mathbf{u}_k$). $\mathbf{U}^{\dagger}$ is a conjugate transpose. $\Lambda$ is a diagonal matrix where $\mathbf{\lambda}_k$ is the $k$-th eigenvalue. The unnormalized Laplacian is also diagonalizable into a similar formula. The magnetic Laplacians represent the topological information of graphs with its eigenvectors and eigenvalues.

\begin{table*}[bt]
\centering
{
\begin{tabular}{ccccc}
\Xhline{2.8\arrayrulewidth}
Laplacian & vanilla & directed & signed & signed directed \\ \hline
$\textbf{L}$(\cite{kipf2016semi}) & \checkmark  & \ding{55} & \ding{55} & \ding{55} \\
$\textbf{L}^q$(\cite{zhang2021magnet}) &  \checkmark & \checkmark & \ding{55} & \ding{55}  \\
\text{proposed} $\textbf{L}^q$ & \checkmark  & \checkmark & \checkmark & \checkmark \\ \Xhline{2.5\arrayrulewidth}
\end{tabular}
}
\caption{Several Laplacians and their applicabilities.}
\label{tab:Dataset Statistics}
\end{table*}

\section{Theoretical Analysis}
\subsection{Encoding Properties}
The proposed complex Hermitian adjacency matrix has two distinctive properties. First, two edges in different direction between a node pair are encoded as a conjugate pair. For example, assume there are two positive edges. One is from node $u$ to $v$, and the other one is from node $v$ to $u$. By the Eq. 2 and Eq. 3, $\textbf{H}^q(u,v)=\frac{1}{2}(\text{cos}q+i\text{sin}q)$ and $\textbf{H}^q(v,u)=\frac{1}{2}(\text{cos}q-i\text{sin}q)$, forming a conjugate pair. 
Thanks to this property, $\textbf{H}^q$ becomes a skew-symmetric Hermitian matrix. 
\begin{equation}
\textbf{H}^{q}(u,v) = \overline{\textbf{H}}^{q}(v,u)
\end{equation}
Second, two edges in opposite signs are encoded as two complex numbers that are summed to be zero. That is the two complex numbers are the negations to each other. Assume there are a positive edge from node $u$ to $v$ and a negative edge from $u$ to $v$. The negative edge is encoded as the negation of the positive edge's. The meanings of edge relations are completely changed by the signs such as (like/dislike, trust/distrust). The proposed edge encoding method reflects the real-world meanings by encoding them in the opposite phase. 
\begin{equation}
    \textbf{H}^{q}(u,v)= 
\begin{cases}
    \ \ \ \frac{1}{2}(\text{cos}q+i\text{sin}q) & \ \ \ \ \ \ \text{if} \ \ \ S(u,v)=1\\
    \ -\frac{1}{2}(\text{cos}q+i\text{sin}q) & \ \ \ \ \ \ \text{if} \ \ \ S(u,v)=-1 \\
    \qquad \qquad 0 &\ \ \ \ \ \ \text{if} \ \ \ S(u,v)=0 \\
\end{cases}
\end{equation}
These properties let the proposed adjacency matrix and magnetic Laplacian satisfy the PSD condition as well as represent node relations without information loss.

\subsection{Generality of the Proposed Laplacian}
The proposed Laplacian can represent any type of graph. For example, edges of undirected unsigned graphs are encoded with \{0,\ 1\} assuming all edges are positive and bidirectional. Then the proposed Laplacian is the same as the traditional graph Laplacian form. Similarly, if a graph is directed and unsigned, its edges are encoded with \{0,\ 1,\ $\frac{1}{2}(\text{cos}q+i\text{sin}q)$,\ $\frac{1}{2}(\text{cos}q-i\text{sin}q)$\}, again assuming all edges are positive. It is the same form of Laplacian defined at MagNet\cite{zhang2021magnet}. In other words, the traditional graph Laplacian and the Laplacian from MagNet\cite{zhang2021magnet} are the special cases of ours. 

\section{Model Framework}
\subsection{Spectral Convolution via the Magnetic Laplacian}
As we discussed in Sec. 4, the magnetic Laplacian ($\mathbf{L}^{q}$) is PSD and is diagonalizable with eigenvector matrix ($\mathbf{U}$) and diagonal eigenvalue matrix ($\Lambda$). Similar to other graph spectral analyses \cite{defferrard2016convolutional,hammond2011wavelets}, we consider the eigenvectors ($\mathbf{u}_k$) as generalizations of discrete Fourier modes. The graph Fourier transform, $\mathbf{x}:V \to \mathbb{C}$, is defined for a signal as $\mathbf{\hat{\textbf{x}}}= \mathbf{U}^{\dagger}\mathbf{x}$. Then we get the inverse Fourier transform formula by the unitarity of the matrix $\mathbf{U}$.
\begin{equation}
 \mathbf{x}=\mathbf{U}\mathbf{\hat{x}}= \sum_{k=1}^{N}\hat{\mathbf{x}}(k)\mathbf{u}_k. 
\end{equation}
The spectral convolution of the graph signal corresponds to the multiplication of signal $\mathbf{x}$ with a filter in the Fourier domain. We defined the spectral convolution as:
\begin{equation}
 \mathbf{g}_\theta \ast \mathbf{x} = \mathbf{U} \mathbf{g}_\theta \mathbf{U}^{\dagger}\mathbf{x},
\end{equation}
where $\mathbf{g}_\theta=diag(\theta)$ is a trainable convolution filter. This filter is a function of eigenvalues of the magnetic Laplacian ($\mathbf{g}_\theta(\Lambda)$). It requires a large computational burden to get eigendecomposition in the case of large graphs. Hammond et al.\cite{hammond2011wavelets} reduced the computational cost of the decomposition by approximating it to a truncated expansion of Chebyshev polynomials as:
\begin{equation}
 \mathbf{g}_{\theta'}(\mathbf{\Lambda})\approx 
 \sum_{k=0}^{K}\theta'_k T_k(\mathbf{\tilde{\Lambda}}),
\end{equation} 
where $T_0(x)=1, T_1(x)=x$, and $T_k=2x T_{k-1}(x)+T_{k-2}(x)$ for $k \geq 2$. $\theta'_k$ are Chebyshev coefficients. $\mathbf{\tilde{\Lambda}}=\frac{2}{\lambda_{max}}\mathbf{\Lambda}-\mathbf{I}$ is a normalized eigenvalue matrix and $\lambda_{max}$ is the largest eigenvalue of the Laplacian. Then the approximated spectral convolution becomes:
\begin{equation}
 \mathbf{g}_{\theta'} \ast \mathbf{x} = 
 \sum_{k=0}^{K}\theta'_k T_k(\mathbf{\tilde{L}})x,
\end{equation} 
where $\mathbf{\tilde{L}}=\frac{2}{\lambda_{max}}\mathbf{L}-\mathbf{I}$ analogous to $\mathbf{\tilde{\Lambda}}$\cite{defferrard2016convolutional,kipf2016semi}. Though the equations look similar to the existing spectral convolutions, the structure of magnetic Laplacian yields a difference. The equation, $\mathbf{g}_{\theta'} \ast \mathbf{x}(u)$, aggregates not only the values of $x$ from the $k$-hop neighborhood of a node $u$ but also the values of $x$ from the nodes that can reach within $k$-hops. The two sets of nodes are the same in case of undirected graphs but not necessarily the same for directed graphs. Moreover, due to the different phases of each relation, the filter ($\mathbf{g}_{\theta'}$) aggregates neighbor information differently. These properties enable our model to aggregate neighbor signals or features effectively and adaptively.

\begin{figure*}[t]
  \begin{center}
  \includegraphics[width=\textwidth]{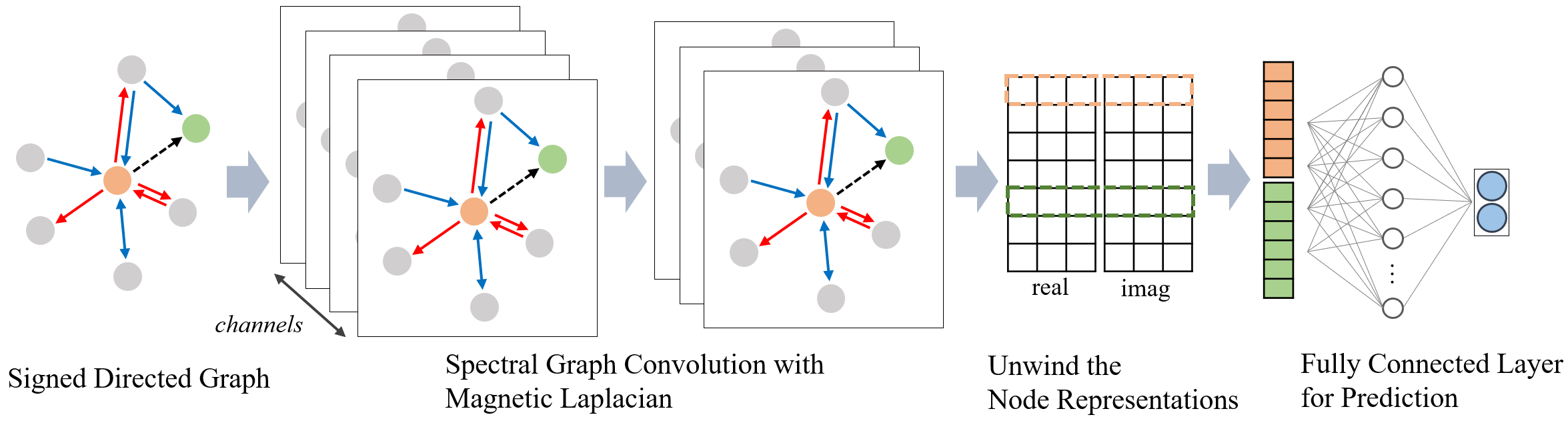}
  \end{center}
  \caption{Overview of the proposed SD-GCN. It shows a link sign prediction example between the orange node and the green node.} 
\end{figure*}

\subsection{Spectral Convolution Layer}
The eigenvalues of the normalized magnetic Laplacian lie in [0, 2] (Please refer to Appendix B.2.). Assuming that $\lambda_{max} \approx 2$ and the maximum polynomial order to be one (i.e. $k=1$). Eq. 13 is simplified to:
\begin{equation}
    \mathbf{g}_{\theta'} \ast \mathbf{x} \approx \theta(\mathbf{I}+(\mathbf{D}^{-\frac{1}{2}}_s \mathbf{A}_s \mathbf{D}^{-\frac{1}{2}}_s) \odot \mathbf{P}^q)\mathbf{x},
\end{equation}
where $\theta=\theta_0'=-\theta_1'$. Then, the spectral convolution is defined as:
\begin{equation}
  (\mathbf{\tilde{D}}^{-\frac{1}{2}}_s \mathbf{\tilde{A}}_s \mathbf{\tilde{D}}^{-\frac{1}{2}}_s\odot \mathbf{P}^q) \mathbf{X}\mathbf{W},
\end{equation} 
where $\mathbf{\tilde{A}_s} = \mathbf{A}_s+\mathbf{I}$ and $\mathbf{\tilde{D}_s}(i,i)=\sum_j{\mathbf{\tilde{A}_s}(i,j)}$. It is a renormalization trick that $\mathbf{I}+(\mathbf{D}^{-\frac{1}{2}}_s \mathbf{A}_s \mathbf{D}^{-\frac{1}{2}}_s)\odot \mathbf{P}^q \to \mathbf{\tilde{D}}^{-\frac{1}{2}}_s \mathbf{\tilde{A}}_s \mathbf{\tilde{D}}^{-\frac{1}{2}}_s\odot \mathbf{P}^q$. It helps prevent the gradient vanishing and exploding issues. 
$\mathbf{X}\in \mathbb{R}^{N \times C}$ is the input signal where $C$ is the input channels. $\textbf{W} \in \mathbb{R}^{C \times F}$ is a learnable weight matrix where $F$ is the number of output channels. 

\subsection{Signed Directed Graph Convolution Network}
SD-GCN stacks $L$ times of the proposed spectral convolutional layer. The $l$-th layer feature vector $\mathbf{x}^{(l)}$ is defined as:
\begin{equation}
  \mathbf{x}_j^{(l)}= \sigma ( \sum_{i=1}^{F_{l-1}} \mathbf{Y}_{ij}^{(l)}\mathbf{x}_i^{(l-1)}+b_j^{(l)}).
\end{equation} 
$\textbf{Y}$ represents the hidden layer of Eq. 15. It expresses the equation in the node-level formula. The bias term is defined as $\mathbf{b}_j^{(l)}(v)=b_j^{(l)}$ and $\text{real}(b_j^{(l)})=\text{imag}(b_j^{(l)})$. The activation function $\sigma$ is a complex version of ReLU\cite{zhang2021magnet}; $\sigma(z)=z$, if $-\pi/2 \leq \text{arg}(z) \leq \pi/2$, and $\sigma(z)=0$, otherwise. $F_{l}$ is the channel number of $l$-th layer. The feature matrix $\mathbf{X}^{(l)}$ has both real- and imaginary-values. At the last convolution layer, we apply unwinding operation:
\begin{equation}
    \mathbf{X}_{\text{unwind}}^{(L)} = 
[\text{real}(\mathbf{X}^{(L)}) \|
    \text{imag}(\mathbf{X}^{(L)}) \otimes (-i)].
\end{equation} 
$\mathbf{X}_{\text{unwind}}^{(L)} \in \mathbb{R}^{N\times 2F_L}$ is unwinded representation of nodes. Finally, we feed them into a fully connected layer:
\begin{equation}
    \textbf{Z} = \sigma(\mathbf{X}_{\text{unwind}}^{(L)}\mathbf{W}_{\text{unwind}} + \mathbf{b}_{\text{unwind}}).
\end{equation}
$\textbf{Z} \in \mathbb{R}^{N\times D}$ is a node embedding matrix and $\textbf{W}_{\text{unwind}} \in \mathbb{R}^{2F_L \times D}$ is a learnable weight matrix where $D$ is the size of node embedding or the number of classes for a prediction task.

\section{Experiments}
We performed extensive experiments to evaluate the performance of the proposed SD-GCN. We adopt the link sign prediction problem as a downstream task, the most common problem in checking the effectiveness of signed directed graph node embedding\cite{huang2021sdgnn,huang2019signed}. The goal of the task is to predict the signs of existing links not observed in the training phase. We adopt the suite for experimental settings popularly used in prior link sign prediction studies\cite{huang2021sdgnn,huang2019signed,li2020learning}. 

\begin{figure*}[t]
\centering
  \includegraphics[width=0.8\linewidth]{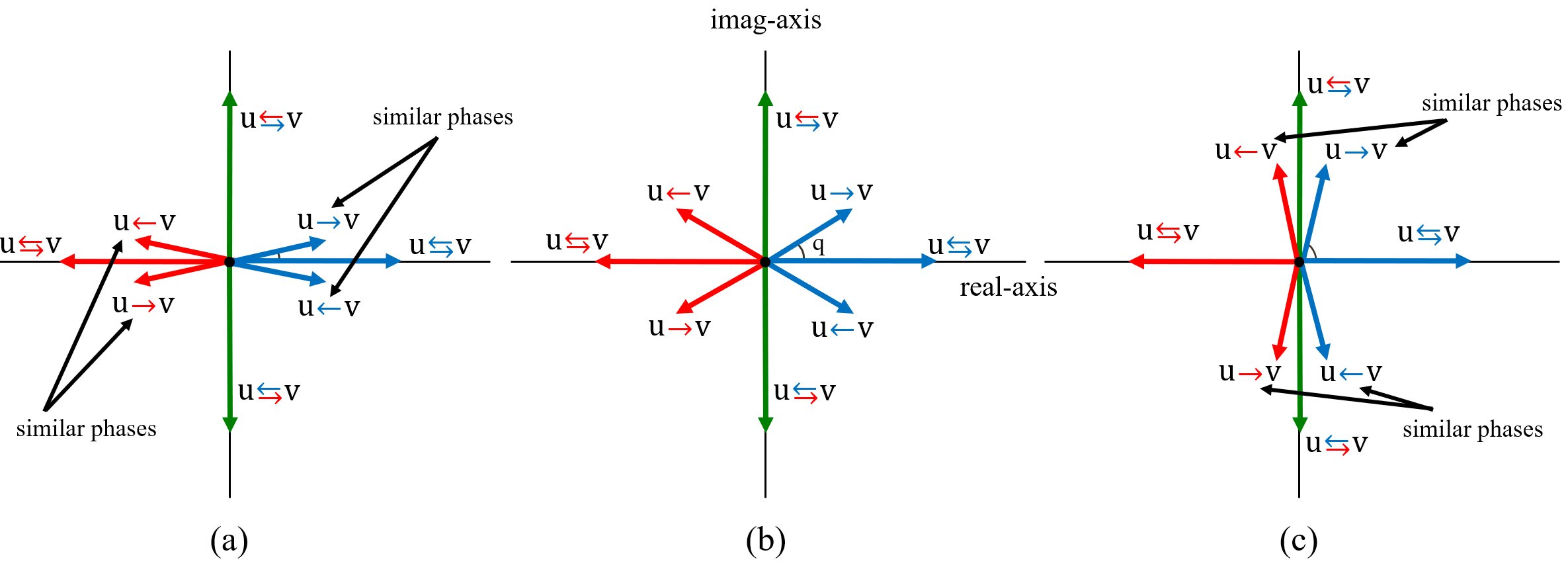}
  \caption{Meaning of $q$ value. (a) shows node relation encoding phases with a small $q$ value. (b) shows node relation encoding phases with a large $q$ value.}
\end{figure*}

\subsection{Datasets and Metrics}
We use four real-world signed directed graph datasets. Bitcoin-Alpha\footnote{
http://www.btc-alpha.com} and Bitcoin-OTC\footnote{http://www.bitcoin-otc.com}\cite{kumar2016edge} are networks from Bitcoin trading platforms. Nodes are users, and the edges indicate who-trust-whom with direction and score. The score is the rate on a scale of -10 to +10. The links with higher than 0 are treated as positive links, and others are considered as negatives. Epinions\footnote{http://www.epinions.com}\cite{guha2004propagation} is another who-trust-whom network from a consumer review site. The edges indicate trust or distrust of reviews written by other users. Slashdot\footnote{http://www.slashdot.com}\cite{kunegis2009slashdot} is a network of user communities from a new website. It tags other users as either friends or foes. 
Table 3 shows the statistics of the four datasets. Node and edge sizes of datasets vary, and positive-negative edge ratios are highly imbalanced. Especially, Bitcoin datasets have more than 90 percent of positive edges, while Slashdot has the smallest positive edge ratio. We adopt four metrics widely used for quantitative comparison, AUC, macro-F1, micro-F1, and binary-F1.

\begin{table*}[h]
\centering
\resizebox{0.65\linewidth}{!}
{
\begin{tabular}{ccccc}
\Xhline{2.5\arrayrulewidth}
Dataset & \# nodes & \# pos links & \# neg links & positive ratio \\ \hline
Bitcoin-Alpha & 3,783 & 22,650 & 1,536 & 0.937 \\
Bitcoin-OTC & 5,881 & 32,029 & 3,563 & 0.900 \\
Epinions & 131,828 & 717,667 & 123,705 & 0.853 \\
Slashdot & 82,140 & 425,072 & 124,130 & 0.774 \\ \Xhline{2.5\arrayrulewidth}
\end{tabular}%
}
\caption{Dataset Statistics.}
\label{tab:Dataset Statistics}
\end{table*}

\subsection{Baseline}
We choose various graph convolution schemes as baselines covering both spatial and spectral methods. 
\begin{itemize}
\item Signed graph: 
\textbf{SGCN}\footnote{https://github.com/benedekrozemberczki/SGCN}\cite{derr2018signed} and \textbf{SNEA}\footnote{https://github.com/liyu1990/snea}\cite{li2020learning} leverage balance theory for signed graph convolution. SGCN defines balanced and unbalanced paths for feature aggregation. SNEA extends SGCN with attention mechanism\cite{bahdanau2014neural}.

\item Directed graph:  \textbf{DiGCN}\footnote{https://github.com/flyingtango/DiGCN}\cite{tong2020digraph} and \textbf{MagNet}\footnote{https://github.com/matthew-hirn/magnet}\cite{zhang2021magnet} define novel Laplacian matrices for directed graphs. DiGCN uses transition and stationary distributions to make a symmetric directed Laplacian. MagNet proposed a complex Hermitian matrix to encode directed edges and define a magnetic Laplacian.

\item Signed Directed graph: \textbf{SiGAT}\cite{huang2019signed} and \textbf{SDGNN}\footnote{https://github.com/huangjunjie-cs/SiGAT}\cite{huang2021sdgnn} leverage not only the balance theory but also the status theory. SiGAT defines triads motifs and uses GAT\cite{velivckovic2017graph}. 
SDGNN defines several weight matrices to learn the different semantics of edge types. 
\end{itemize}

\subsection{Implementation details}
We ignore the sign information of the datasets when training directed graph models and utilize direction information only. On the contrary, SGCN and SNEA use sign information only and ignore the directions. We set the node embedding dimension to 64 for all the baselines to provide the same learning capacity. We use the same hyper-parameter settings adopted in each baseline. For our model, we set $q=0.1\pi$ and stack two spectral convolution layers ($L=2$). For the link sign prediction task, we concatenate the two unwinding values of node pairs and feed them into a fully connected layer with the softmax function (see Fig. 3). We use the negative log-likelihood loss function and Adam optimizer with learning rate = $1e^{-3}$, weight decay = $1e^{-5}$. The links are split into 60:20:20 for training, validation, and test sets. Positive and negative edges are sampled at a ratio of 3:1 in the training phase. It is because the ratio of the positive and negative edges is unbalanced. If we train with 90 percent of positive samples, a model can easily improve its performance by simply predicting all links are positive. Fig. 6 shows the effect of varying positive sample ratios. The performances are the average score of 10 experiments from different seed sets. (The statistical null hypothesis test results are listed in Appendix A.) The experiments are conducted on the environment of Ubuntu v16.4 with Xeon E5-2660 v4 and Nvidia Titan XP 12G. It is implemented via python v3.6 and Pytorch v1.8.0.

\section{Results}

\begin{table*}[t]
\centering
\resizebox{0.9\textwidth}{!}{%
\begin{tabular}{cc|cc|cc|ccc}
\Xhline{2\arrayrulewidth}
&  & \multicolumn{2}{c|}{Signed} & \multicolumn{2}{c|}{Directed} & \multicolumn{3}{c}{ Signed Directed}   \\ \hline
\multicolumn{1}{c|}{Dataset} & Metric & SGCN & SNEA & DiGCN & MagNet & SiGAT & SDGNN &  SD-GCN \\   \Xhline{2\arrayrulewidth}
\multicolumn{1}{c|}{\multirow{4}{*}{Bitcoin-Alpha}} & AUC & 0.779 & 0.812 & 0.823 & 0.840 & 0.844 & \underline{0.847} &  \textbf{0.887} \\
\multicolumn{1}{c|}{} & Macro-F1  & 0.662 & 0.668 & 0.681 & \underline{0.702} &  0.671 & 0.682 & \textbf{0.750}  \\
\multicolumn{1}{c|}{} & Micro-F1  & 0.914 & 0.860 & 0.902 & \underline{0.931} & 0.896 & 0.905  & \textbf{0.935} \\
\multicolumn{1}{c|}{} & Binary-F1  & 0.954 & 0.920 & 0.935 & \underline{0.963} & 0.943 & 0.948 &  \textbf{0.965}\\  \hline
\multicolumn{1}{c|}{\multirow{4}{*}{Bitcoin-OTC}} & AUC & 0.834 & 0.830 & 0.879 & 0.874 & 0.878 & \underline{0.889} & \textbf{0.917} \\
\multicolumn{1}{c|}{} & Macro-F1  & 0.704 & 0.742 & 0.739 &  0.746 & 0.745 & \underline{0.757} & \textbf{0.809}  \\
\multicolumn{1}{c|}{} & Micro-F1  & 0.897 & 0.871 & 0.896 & \underline{0.921} & 0.901 & 0.908 & \textbf{0.929} \\
\multicolumn{1}{c|}{} & Binary-F1 & 0.943 & 0.924 &  0.940 & \underline{0.957} & 0.945 & 0.949 & \textbf{0.961} \\  \hline
\multicolumn{1}{c|}{\multirow{4}{*}{Epinions}} & AUC & 0.843 & 0.829 & 0.874  &  0.912 & 0.904 & \underline{0.913} & \textbf{0.941} \\
\multicolumn{1}{c|}{} & Macro-F1  & 0.744 & 0.783 & 0.781 & 0.825 & 0.800 & \underline{0.837} &  \textbf{0.862} \\
\multicolumn{1}{c|}{} & Micro-F1  & 0.899 & 0.872 & 0.888 & \underline{0.923} & 0.900 & 0.921 & \textbf{0.934} \\
\multicolumn{1}{c|}{} & Binary-F1 & 0.943 & 0.922 & 0.942 & \underline{0.956} & 0.941 & 0.954 & \textbf{0.962} \\  \hline
    \multicolumn{1}{c|}{\multirow{4}{*}{Slashdot}} & AUC & 0.732 & 0.794 &  0.754 & 0.765 & 0.843 & \underline{0.849} & \textbf{0.893} \\
\multicolumn{1}{c|}{} & Macro-F1  & 0.559 & 0.764 & 0.589 & 0.575 & 0.722 & \underline{0.731} &  \textbf{0.786} \\
\multicolumn{1}{c|}{} & Micro-F1  & 0.785 & 0.817 & 0.803 &  0.793 & 0.822 & \underline{0.826} & \textbf{0.858} \\
\multicolumn{1}{c|}{} & Binary-F1  & 0.875 & 0.876  & 0.879 & 0.880 & 0.889 & \underline{0.891} & \textbf{0.910} \\   \Xhline{2\arrayrulewidth}
\end{tabular}%
}
\caption{Link sign prediction performance. \textbf{Bold} indicates the best performance, and \underline{underline} tells the second best. The performances are the average of ten experiments from different seed sets.}
\label{tab:my-table}
\end{table*}

\subsection{Link Sign Prediction}
Table 4 shows the link sign prediction results. The baselines are classified into three categories; signed model, directed model, and signed directed model. 
First, we can observe that the proposed SD-GCN outperforms all baselines for all datasets and in four performance metrics. SD-GCN can be considered as an advanced model of MagNet\cite{zhang2021magnet}, equipped with a novel sign encoding mechanism. Therefore, we may expect that SD-GCN provides better performance than MagNet, and its advantage is more distinctive when a dataset has a more balanced positive-negative edge ratio. Our experimental results confirm the expectation. The performance enhancement of SD-GCN over MagNet gets larger in Epinions and Slashdot than in the two Bitcoin datasets where positive edges overwhelm negative edges. Even though MagNet and SD-GCN have similar micro-F1 and binary-F1 scores in Bitcoin-Alpha and OTC, SD-GCN always shows the best performance regardless of the positive-negative ratio or of the network size. The performance results show that the proposed magnetic Laplacian embeds both sign and direction information appropriately in representation learning task. 
MagNet, a directed spectral model, and SDGNN\cite{huang2021sdgnn}, a signed directed spatial model, show the second-best performance depending on the properties of datasets. In the case of Bitcoin-Alpha which has the highest positive-negative edge ratio, sign information may have reduced discriminative power compared to direction information. Therefore, MagNet, a directed model, performs well on Bitcoin-Alpha and shows the second-best performance. On the contrary, on the Slashdot dataset with the smallest positive-negative edge ratio, sign information gains its discriminative power. Therefore, SDGNN, a signed directed model, enjoys the second-best performance on Slashdot. Our SD-GCN considers both sign and direction information and always performs better than MagNet and SDGNN.

\subsection{$q$ value analysis}
The proposed complex Hermitian adjacency matrix requires the parameter value $q$ between [0, $\pi/2$]. The node relationships are encoded with sine and cosine functions as shown in Eq. 2 and Eq. 3, and the parameter $q$ plays an important role in encoding. Fig. 5 shows the sign prediction performance as a function of $q$. Though the performance is not very sensitive to $q$, we can observe peculiar patterns in performance changes. For example, all four performance metrics decrease when $q$ is more than or equal to 0.3$\pi$. Fig. 4 illustrates the effect of $q$ on edge encoding. When the $q$ is small, the phase difference between two edges in opposite directions is small (Fig. 4(a)). In an extreme case of $q=0$, edge direction information is ignored and the proposed scheme becomes an undirected model. Fig. 4(b) shows a case of large $q$. In this case, two edges with inverse direction and inverse sign are encoded with similar phases. It detriments not only edge encoding quality but also influence the prediction performance as illustrated in Fig. 5. 

\subsection{Sampling Ratio}
As 70-90\% of edges are positive, we control the number of positive edge samples during the training stage to improve training efficiency and performance. Fig. 6 shows the performance as a function of the positive edge sampling ratio. The best performances are achieved when the ratio is between 3 to 5. Low performances at low ratios indicate the risk of excessive under-representation of positive edges.

\begin{figure}[t]
  \centering
  \includegraphics[width=\columnwidth]{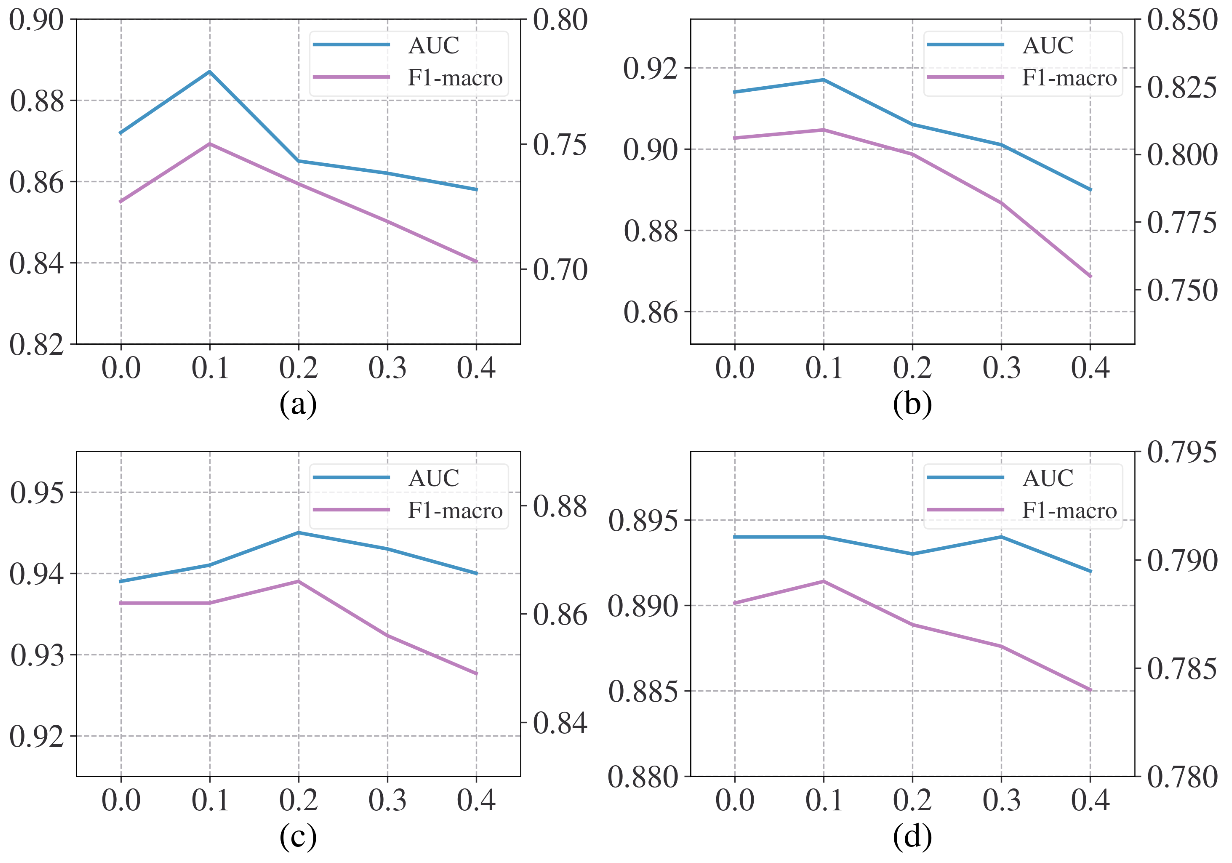}
  \caption{Performance variation by the $q$ value. The scale of the x-axis is $\pi$. (a), (b), (c), and (d) are the results of Bitcoin-Alpha and Bitcoin-OTC, Epinions, and Slashdot, respectively.}
\end{figure}

\begin{figure}[t]
  \centering
  \includegraphics[width=\columnwidth]{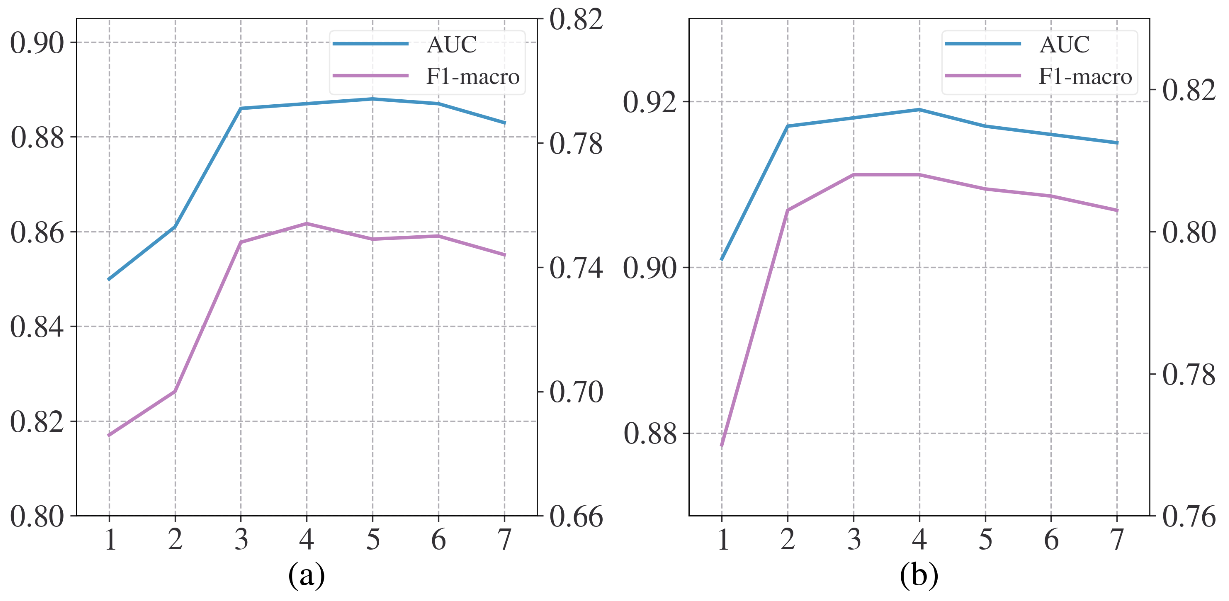}
  \caption{Performance variation by the positive sampling ratio. (a) and (b) are for Bitcoin-Alpha and Bitcoin-OTC, respectively.}
\end{figure}

\section{Conclusion}
In this paper, we propose a spectral convolution model called SD-GCN for signed directed graph analyses. We introduce a novel complex Hermitian adjacency matrix that can encode both edge sign and direction information. Based on the Hermitian, we define the magnetic Laplacian matrix. The Laplacian matrix is positive semi-definite such that we apply the graph spectral convolution method for its spectral analyses. SD-GCN architecture with the convolution layers is derived from the magnetic Laplacian. To the best of our knowledge, it is the first spectral model for signed graphs. Extensive experiments were performed on four real-world datasets to compare the performance of the proposed method with the state-of-the-art baselines. The experiments show that the proposed SD-GCN outperforms the other graph convolutions.

\bibliographystyle{aaai}
\bibliography{reference}

\onecolumn

\appendix
\section*{APPENDIX}
\section{Experiment Details}
\subsection{Data Preprocessing}
The preprocessed datasets can be found at  Standford Network Analysis Project (SNAP)\footnote{https://snap.stanford.edu/data/index.html\#signnets}.
Some papers\cite{li2020learning,derr2018signed} used sub-networks of the originals due to the large network size. We use the whole graph structure for the experiments. As discussed earlier, each positive and negative edge sets are split into three sets, training, validation, and test. Then the positive and negative training edge sets compose the training set. It is the same for validation and test sets.

\subsection{Training}
We ran ten times of experiments with different seed sets for a fair comparison. The seeds are [0, 10, 20, 30, 40, 50, 60, 70, 80, 90]. We apply early stop conditions by comparing the training and validation losses. The model parameters with the lowest validation loss are saved during the training. If validation loss goes up consecutively for more than ten epochs, we stop training and get performance with test set. 

Table A.2 shows the details of the experimental results. The parenthesized values show the standard deviation of the experiments. We do not conclude the excellency of our model only with the average values. We apply Weltch's unequal variances t-test\cite{welch1947generalization} for the statistically proven conclusion. The null hypothesis of the test is that the performance of SD-GCN is higher than the second-best. The last column of Table A1 shows the p-value significance. *, **, and *** are the significant level of 0.1, 0.05, and 0.01, respectively. Except for one experiment, we conclude that the proposed model shows higher performance than the others with strong confidence.
\renewcommand{\thetable}{\thesubsection}
\begin{table}[h]
\centering
\resizebox{0.9\textwidth}{!}
{\begin{tabular}{cc|c|cc|c}
\Xhline{2\arrayrulewidth}
&  &  \multicolumn{1}{c|}{Directed} & \multicolumn{2}{c|}{ Signed Directed} &   \\ \hline
\multicolumn{1}{c|}{Dataset} & Metric & MagNet & SDGNN &  SD-GCN & significance \\   \Xhline{2\arrayrulewidth}
\multicolumn{1}{c|}{\multirow{4}{*}{Bitcoin-Alpha}} & AUC & 0.840\small(0.0197) & \underline{0.847}\small(0.0157) &  \textbf{0.887}\small(0.0221) & ***\\
\multicolumn{1}{c|}{} & Macro-F1  &  \underline{0.702}\small(0.0221)  &  0.682\small(0.0142) & \textbf{0.750}\small(0.0243) & ***  \\
\multicolumn{1}{c|}{} & Micro-F1  &  \underline{0.931}\small(0.0043) &  0.905\small(0.0057)  & \textbf{0.935}\small(0.0022) & ***\\
\multicolumn{1}{c|}{} & Binary-F1  & \underline{0.963}\small(0.0022) &  0.948\small(0.0032) &  \textbf{0.965}\small(0.0040)& *\\  \hline
\multicolumn{1}{c|}{\multirow{4}{*}{Bitcoin-OTC}} & AUC & 0.874\small(0.0113) & \underline{0.889}\small(0.0085) & \textbf{0.917}\small(0.0077)& *** \\
\multicolumn{1}{c|}{} & Macro-F1  &  0.746\small(0.0099) & \underline{0.757}\small(0.0157) & \textbf{0.809}\small(0.0065) & *** \\
\multicolumn{1}{c|}{} & Micro-F1  &  \underline{0.921}\small(0.0025) & 0.908\small(0.0066) & \textbf{0.929}\small(0.0034) & ***\\
\multicolumn{1}{c|}{} & Binary-F1 &  \underline{0.957}\small(0.0015) & 0.949\small(0.0037) & \textbf{0.961}\small(0.0018) & ***\\  \hline
\multicolumn{1}{c|}{\multirow{4}{*}{Epinions}} & AUC & 0.912\small(0.0062) & \underline{0.913}\small(0.0035) & \textbf{0.941}\small(0.0013)& *** \\
\multicolumn{1}{c|}{} & Macro-F1  &  0.825\small(0.0030) & \underline{0.837}\small(0.0032) &  \textbf{0.862}\small(0.0026)& *** \\
\multicolumn{1}{c|}{} & Micro-F1  &  \underline{0.923}\small(0.0031) & 0.921\small(0.0015) & \textbf{0.934}\small(0.0014) & ***\\
\multicolumn{1}{c|}{} & Binary-F1 &  \underline{0.956}\small(0.0005) & 0.954\small(0.0009) & \textbf{0.962}\small(0.0010) & ***\\  \hline
    \multicolumn{1}{c|}{\multirow{4}{*}{Slashdot}} & AUC & 0.765\small(0.0043) & \underline{0.849}\small(0.0029) & \textbf{0.893}\small(0.0014) & ***\\
\multicolumn{1}{c|}{} & Macro-F1   & 0.575\small(0.0110) & \underline{0.731}\small(0.0037) &  \textbf{0.786}\small(0.0018) & ***\\
\multicolumn{1}{c|}{} & Micro-F1   &  0.793\small(0.0023)  & \underline{0.826}\small(0.0016) & \textbf{0.858}\small(0.0011) & ***\\
\multicolumn{1}{c|}{} & Binary-F1 & 0.880\small(0.0010) & \underline{0.891}\small(0.0007) & \textbf{0.910}\small(0.0007) & ***\\   \Xhline{2\arrayrulewidth}
\end{tabular}
}
\caption{Significant Test.}
\label{a_table}
\end{table}

\section{Proof of Theorems}
In this section, we proved the theorems omitted in the main text. First, we prove Theorem 1 that magnetic Laplacian is a positive semi-definite (PSD). Then, we show Theorem 2 that the eigenvalues of the normalized magnetic Laplacian are between 0 to 2. These theorems are the key factors for deriving the proposed spectral convolution. 

\subsection{Positive Semi-definite of the proposed Magnetic Laplacian}
\begin{theorem} For a signed directed graph $\mathcal{G} = (V, E, \textbf{\textit{S}})$, both the unnormalized and normalized magentic Laplacian $\textbf{L}_{U}^{q}, \textbf{L}_{N}^{q}$ are positive semdifinite.
\end{theorem}
\textit{proof.} \\ The unnormalized magnetic Laplacian $\textbf{L}_{U}^{q}$ is an Hermitian matrix by its definition. Then, we have Imag($\textbf{x}^{\dagger}\textbf{L}_{U}^{q}\textbf{x}$)=0 where {$\textbf{x} \in \mathbb{C}^N $}. Now we show Real($\textbf{x}^{\dagger}\textbf{L}_{U}^{q}\textbf{x}$) $\geq 0$. The following procedures utilize the definitions of $\text{D}_s$ and $\text{A}_s$.
\begin{flalign*}
& 2\text{Real}(\textbf{x}^{\dagger}\textbf{L}_{U}^{q}\textbf{x}) \\
= & 2\sum_{u,v=1}^{N}{\mathbf{D_s}}(u,v)\mathbf{x}(u)\overline{\mathbf{x}(v)}\\ & - 2\sum_{u,v=1}^{N}{\mathbf{A_s}}(u,v)\mathbf{x}(u)\overline{\mathbf{x}(v)}
\left[
    \frac{\text{cos}(i\Theta_{uv}^{q})\text{A}_{uv}+ \text{cos}(i\overline{\Theta}^{q}_{uv})\text{A}_{vu}}
    {\|\text{exp}(i\Theta_{uv}^{q})\text{A}_{uv}+ \text{exp}(i\overline{\Theta}^{q}_{uv})\text{A}_{vu}\|+ \epsilon}
\right]\\
= & 2\sum_{u=1}^{N}{\mathbf{D_s}}(u,u)\mathbf{x}(u)\overline{\mathbf{x}(u)}\\ & - 2\sum_{u,v=1}^{N}{\mathbf{A_s}}(u,v)\mathbf{x}(u)\overline{\mathbf{x}(v)}
\left[
     \frac{\text{cos}(i\Theta_{uv}^{q})\text{A}_{uv}+ \text{cos}(i\overline{\Theta}^{q}_{uv})\text{A}_{vu}}
    {\|\text{exp}(i\Theta_{uv}^{q})\text{A}_{uv}+ \text{exp}(i\overline{\Theta}^{q}_{uv})\text{A}_{vu}\|+ \epsilon}
    \right]\\
= & 2\sum_{u,v=1}^{N}{\mathbf{A_s}}(u,v)\abs{\mathbf{x}(u)}^2\\ & - 2\sum_{u,v=1}^{N}{\mathbf{A_s}}(u,v)\mathbf{x}(u)\overline{\mathbf{x}(v)}
\left[
     \frac{\text{cos}(i\Theta_{uv}^{q})\text{A}_{uv}+ \text{cos}(i\overline{\Theta}^{q}_{uv})\text{A}_{vu}}
    {\|\text{exp}(i\Theta_{uv}^{q})\text{A}_{uv}+ \text{exp}(i\overline{\Theta}^{q}_{uv})\text{A}_{vu}\|+ \epsilon}
\right]\\
= & \sum_{u,v=1}^{N}{\mathbf{A_s}}(u,v)\abs{\mathbf{x}(u)}^2 + \sum_{u,v=1}^{N}{\mathbf{A_s}}(u,v)\abs{\mathbf{x}(v)}^2\\ & - 2\sum_{u,v=1}^{N}{\mathbf{A_s}}(u,v)\mathbf{x}(u)\overline{\mathbf{x}(v)}
\left[
     \frac{\text{cos}(i\Theta_{uv}^{q})\text{A}_{uv}+ \text{cos}(i\overline{\Theta}^{q}_{uv})\text{A}_{vu}}
    {\|\text{exp}(i\Theta_{uv}^{q})\text{A}_{uv}+ \text{exp}(i\overline{\Theta}^{q}_{uv})\text{A}_{vu}\|+ \epsilon}
\right]\\
= & \sum_{u,v=1}^{N}{\mathbf{A_s}}(u,v)
\left(
\abs{\mathbf{x}(u)}^2 +\abs{\mathbf{x}(v)}^2 -
2\mathbf{x}(u)\overline{\mathbf{x}(v)}
\left[
     \frac{\text{cos}(i\Theta_{uv}^{q})\text{A}_{uv}+ \text{cos}(i\overline{\Theta}^{q}_{uv})\text{A}_{vu}}
    {\|\text{exp}(i\Theta_{uv}^{q})\text{A}_{uv}+ \text{exp}(i\overline{\Theta}^{q}_{uv})\text{A}_{vu}\|+ \epsilon}
\right]
\right) \\
\geq & \sum_{u,v=1}^{N}{\mathbf{A_s}}(u,v) 
(\abs{\mathbf{x}(u)}^2 +\abs{\mathbf{x}(v)}^2 -
2\abs{\mathbf{x}(u)}\abs{\overline{\mathbf{x}(v)}})\\
= & \sum_{u,v=1}^{N}{\mathbf{A_s}}(u,v) 
(\abs{\mathbf{x}(u)}-\abs{\mathbf{x}(v)})^2\\
\geq & 0.
\end{flalign*}

Thus, $\textbf{x}^{\dagger}\textbf{L}_{U}^{q}\textbf{x} \geq 0$ for $\textbf{x} \in \mathbb{C}^N $, positive semi-definite. \\
For normalized Laplacian matrix, $\textbf{L}_{N}^{q} = \textbf{D}_{s}^{-1/2}\textbf{L}_{U}^{q}\textbf{D}_{s}^{-1/2}$.
\begin{flalign*}
\textbf{x}^\dagger\textbf{L}_{N}^{q}\textbf{x} 
& = \textbf{x}^\dagger\textbf{D}_{s}^{-1/2}\textbf{L}_{U}^{q}\textbf{D}_{s}^{-1/2}\textbf{x} \\
& =\textbf{y}^\dagger\textbf{L}_{U}^{q}\textbf{y} \\
& \geq 0.
\end{flalign*} 
where, $\textbf{y}=\textbf{D}_{s}^{-1/2}\textbf{x}$. Thus, both unnormalized and normalized magnetic Laplacians are  positive semi-definite. 

\subsection{Interval of the Normalized Magnetic Laplacian Eigenvalues}
\begin{theorem} For a signed directed graph $\mathcal{G} = (V, E, \textbf{\textit{S}})$, 
the eigenvalues of the normalized magnetic Laplacian $\textbf{L}_{N}^{q}$ lie in [0, 2].
\end{theorem}
\textit{proof.} \\
$\textbf{L}_{N}^{q}$ has non-negative and real eigenvalues since it is positive semi-definite by Theorem.1. Now, we show the eigenvalues are less than or equal to 2. Here, we use the Courant-Fischer theorem\cite{golub2013matrix},
\begin{flalign*}
\lambda_{N} 
& = \max_{\textbf{x}\neq0} \frac{\textbf{x}^{\dagger}\textbf{L}_{N}^{q}\textbf{x}}{\textbf{x}^{\dagger}\textbf{x}} \\
& = \max_{\textbf{x}\neq0} \frac{\textbf{x}^{\dagger}\textbf{D}_s^{-1/2}\textbf{L}_{U}^{q}\textbf{D}_s^{-1/2}\textbf{x}}{\textbf{x}^{\dagger}\textbf{x}} \\
& = \max_{\textbf{y}\neq0} \frac{\textbf{y}^{\dagger}\textbf{L}_{U}^{q}\textbf{y}}{\textbf{y}^{\dagger}\textbf{D}_{s}\textbf{y}}.
\end{flalign*} 
where, $\textbf{y}=\textbf{D}_{s}^{-1/2}\textbf{x}$. Since $\textbf{D}_s$ is diagonal, 
\begin{equation*}
\textbf{y}^\dagger\textbf{D}_s\textbf{y}=\sum_{u,v=1}^{N}{\textbf{D}_s(u,v)}\textbf{y}(u)\overline{\textbf{y}(v)} = \sum_{u=1}^{N}{\textbf{D}_s(u,u)\abs{\textbf{y}}^2}
\end{equation*}
Similar to Theorem 1, we have
\begin{flalign*}
&\textbf{y}^\dagger\textbf{L}_U^q\textbf{y}  \\
= & \frac{1}{2}\sum_{u,v=1}^N{\textbf{A}_s}(u,v)
    \left(\abs{\textbf{y}(u)}^2+\abs{\textbf{y}(v)}^2-2\textbf{y}(u)\overline{\textbf{y}(v)}
     \frac{\text{cos}(i\Theta_{uv}^{q})\text{A}_{uv}+ \text{cos}(i\overline{\Theta}^{q}_{uv})\text{A}_{vu}}
    {\|\text{exp}(i\Theta_{uv}^{q})\text{A}_{uv}+ \text{exp}(i\overline{\Theta}^{q}_{uv})\text{A}_{vu}\|+ \epsilon}
    \right)\\
\leq & \frac{1}{2}\sum_{u,v=1}^N{\textbf{A}_s}(u,v)
    (\abs{\textbf{y}(u)}^2+\abs{\textbf{y}(v)}^2) \\
\leq & \sum_{u,v=1}^N{\textbf{A}_s}(u,v)
    (\abs{\textbf{y}(u)}^2+\abs{\textbf{y}(v)}^2) \\
\leq & 2\sum_{u,v=1}^N{\textbf{A}_s}(u,v)\abs{\textbf{y}(u)}^2 \qquad  \text{(since}\; \textbf{A}_s \;\text{is symmetric)}\\
= & 2\sum_{u=1}^N{\abs{\textbf{y}(u)}^2}
    \left( \sum_{v=1}^{N}{\textbf{A}_s}(u,v)\right)\\
= & 2\sum_{u=1}^N{\abs{\textbf{y}(u)}^2\textbf{D}_s(u,u)} \\
= & 2\textbf{y}^{\dagger}\textbf{D}_s\textbf{y}.
\end{flalign*}
Thus,
\begin{equation*}
    \lambda_N = 
     \max_{\textbf{y}\neq0} \frac{\textbf{y}^{\dagger}\textbf{L}_{U}^{q}\textbf{y}}
     {\textbf{y}^{\dagger}\textbf{D}_{s}\textbf{y}} \leq
     \max_{\textbf{y}\neq0} \frac{2\textbf{y}^{\dagger}\textbf{D}_s\textbf{y}}
     {\textbf{y}^{\dagger}\textbf{D}_{s}\textbf{y}} = 2.
\end{equation*}
Finally, the eigenvalues of normalized magnetic Laplacian are between [0, 2].

\end{document}